\documentclass[11pt,a4paper]{article}
\usepackage[hyperref]{acl2021}
\usepackage{times}
\usepackage{latexsym}
\usepackage{xspace}
\usepackage{booktabs}
\usepackage{multirow}
\usepackage{makecell}
\usepackage{color}
\usepackage{microtype}
\usepackage{amsmath}
\usepackage{graphicx}
\usepackage{amssymb,amsthm,enumitem}
\usepackage{xcolor}
\usepackage{subcaption}
\usepackage{caption}
\definecolor{applegreen}{rgb}{0.0, 0.42, 0.24}

\newcommand\smallsection[1]{\noindent\textbf{#1.}}
\newcommand{\mtt}[1]{$\mathtt{#1}$}
\newcounter{example}[section]
\usepackage{makecell}
\newenvironment{example}[1][]{\refstepcounter{example}\par\smallskip
   \textbf{Example~\theexample. #1} \rmfamily}{\smallskip}

\aclfinalcopy 

\newcommand{\our}{\mbox{  \textsc{TaLLoR}  }\xspace}

\title{Weakly Supervised Named Entity Tagging with Learnable Logical Rules}

\author{Jiacheng Li$^1$\thanks{~ Work done during an internship at Bosch Research.}, Haibo Ding$^2$, Jingbo Shang$^1$, Julian McAuley$^1$, Zhe Feng$^2$\\
  University of California, San Diego$^1$ \\
  Bosch Research North America$^2$ \\
  \texttt{\{j9li,jshang,jmcauley\}@eng.ucsd.edu}$^1$\\
  \texttt{\{haibo.ding, zhe.feng2\}@us.bosch.com}$^2$\\
}
\date{}

\begin{document}
\maketitle
\begin{abstract}
We study the problem of building entity tagging systems by using a few rules as weak supervision.
Previous methods mostly focus on disambiguating 
entity types based on contexts and 
expert-provided rules,
while assuming entity spans are given.
In this work, we propose a novel method \our that bootstraps high-quality logical rules to train a neural tagger in a fully automated manner.
Specifically, we introduce compound rules that are composed from simple rules to increase the precision of boundary detection and generate more diverse pseudo labels.
We further design a dynamic label selection strategy to ensure pseudo label quality and therefore avoid overfitting the neural tagger.
Experiments on three datasets demonstrate that our method outperforms other weakly supervised methods and even rivals 
a state-of-the-art distantly supervised tagger with a lexicon of over 2,000 terms when starting from only 20 simple rules.
Our method 
can serve as a
tool for rapidly building taggers in emerging domains and tasks.
Case studies show that learned rules can potentially explain the predicted entities.

\end{abstract}

\section{Introduction}

Entity tagging systems 
that follow
supervised training, 
while
accurate, often 
require
a large amount of manual, domain-specific labels, 
making them
difficult to 
apply to
emerging domains and tasks. To reduce manual effort, previous works resort to manual lexicons~\cite{ Shang2018LearningNE, Peng2019DistantlySN} or heuristic rules provided by domain experts~\cite{Fries2017SwellSharkAG,Safranchik2020WeaklySS,lison-etal-2020-named}
as weak supervision.
For example, LinkedHMM~\cite{Safranchik2020WeaklySS} can achieve 
performance close to supervised models using 186 heuristic rules in addition to a lexicon of over two million terms.  
However, it is challenging for experts to write complete and accurate rules or lexicons in emerging domains, which requires both a significant amount of manual 
effort
and a deep understanding of 
the target data.
How to build accurate entity tagging systems using less manual effort is still an open problem.

\begin{figure}[t]
	\centering
	\includegraphics[width=\linewidth]{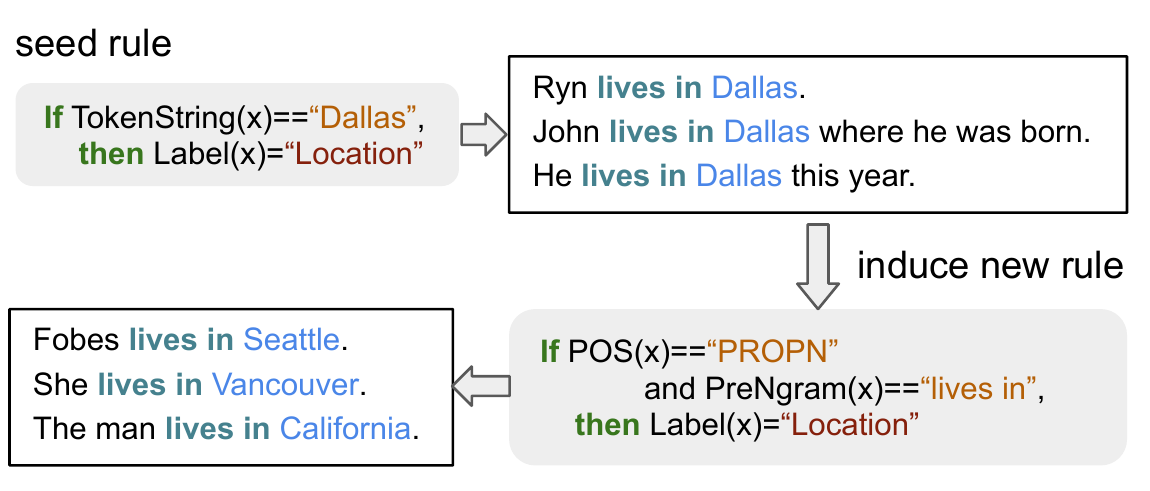}
    \caption{
    Examples of a seed logical rule and a newly induced rule from labeled data for recognizing locations. `x' denotes a token span from a given sentence. 
    }
    \label{fig:logicalrule}
\end{figure}

In this work, we explore methods that can automatically learn new rules from unlabeled data and a small set of seed rules (e.g.~20 rules). 
Such methods are desirable in real-world applications not only because
they
can be rapidly deployed to new domains or customized entity types,
but also
because the learned rules are often effective, interpretable, and simple for non-experts
to ``debug'' incorrect predictions.
As explained in Figure~\ref{fig:logicalrule}, new rules can be learned from seed rules. 
Specifically, we propose a novel iterative learning method \our that can learn accurate rules to train a neural tagger in an 
automated manner, 
with goal to address two key issues during learning process:
(1) how to detect entity boundaries and predict their types simultaneously with rules,
(2) how to generate accurate and diverse pseudo labels from rules.


With such a small set of seed rules as supervision,
previous works~\cite{niu2003bootstrapping,Huang2010InducingDS,Gupta2014ImprovedPL} only focus on disambiguating entity types assuming entity spans are given or just syntactic chunks (e.g.,~noun phrases). However, we find that syntactic chunks often do not align well with target entity spans. 
For example, given a sentence from CoNLL2003: ``{\it Germany's representative to the European Union's veterinary committee...}'', the noun phrases\footnote{Noun phrases are extracted using spaCy noun chunks.} are ``{\it Germany's representative}'' and ``{\it the European Union's veterinary committee}'', but gold entities in the sentence are ``{\it Germany}'' and ``{\it European Union}''.
We used noun phrases extracted from spaCy 
as predicted entity boundaries and compared them with ground truth entity boundaries, which are extracted based on the results from syntactic parsing. This setting of using noun phrases as entity candidates is similar to
previous work~\cite{niu2003bootstrapping, Huang2010InducingDS}. The results are
shown in Table~\ref{tab:bound_detection}, a majority of target entities are missed if we use noun phrases as entity candidates, which will not be recognized correctly later.

To address both entity boundary detection and type classification simultaneously, 
we first define five types of simple logical rules considering 
the lexical, local context, and syntax information of entities. 
We notice that simple logical rules are often inaccurate when detecting entity boundaries. 
Therefore, we propose to learn compound logical rules, which are composed from multiple simple rules and logical connectives (e.g.~``and''). 
For example, given the sentence ``{\it John lives in Dallas where he was born}'', the simple rule ``lives in \underline{~~}'', which is a preceding context clue, will match multiple token spans such as ``{\it Dallas}'', ``{\it Dallas where}'', ``{\it Dallas where he}'' etc. 
In contrast, compound logical rules can both detect entity boundaries  and classify their types accurately.
For example, using both the preceding context and the part-of-speech (POS) tag rule (e.g.~``lives in \underline{~~}'' and POS is a proper noun) can correctly identify the Location entity ``Dallas''.

\begin{table}[t]
    \center
    \small
    \begin{tabular}{r|ccc|ccc}
        \toprule
                & \multicolumn{3}{c}{Noun phrase}  & \multicolumn{3}{c}{\our} \\
                    
        \cmidrule{2-7}
             &  P & R & F$_1$ & P & R & F$_1$ \\
        \midrule
        
        BC5CDR & 17.1 & 50.1 & 25.5 & 69.8 & 67.8 & 68.7\\
        CHEM & 3.2 & 35.6 & 5.8 & 63.0 & 60.2 & 61.6\\
        CoNLL & 4.1 & 47.3 & 7.5 & 86.9 & 86.7 & 86.8\\
        
        \bottomrule
    \end{tabular}
    \caption{
   Boundary detection performance from our method and parsing based noun phrases.
  }
    \label{tab:bound_detection}
    \end{table}

Though we aim to learn accurate rules, automatically acquired rules can be noisy. 
To ensure the quality of generated pseudo labels, we design a dynamic label selection strategy to select highly accurate labels so that the neural tagger can learn new entities instead of overfitting to the seed rules. 
Specifically, we maintain a high-precision label set during our learning process. 
For each learning iteration, we first automatically estimate a filtering threshold based on the high-precision set. 
Then, we filter out low-confidence pseudo labels by considering both 
their
maximum and average distances to the high-precision set. 
Highly confident labels are added into the high-precision set for the next iteration of learning. 
Our dynamic selection strategy enables our framework to maintain the precision of recognized entities while increasing recall during the learning process, as shown in our experiments.

We evaluate our method on three datasets. 
Experimental results show that \our
outperforms existing weakly supervised methods and 
can increase 
the
average $F_1$ score by $60\%$
across three datasets over methods using seed rules. 
Further analysis shows that \our can achieve similar performance with a state-of-the-art distantly supervised method trained using 
1\% of the
human effort\footnote{In experiments, our method used 20 rules, the other system used a manually constructed lexicon of over 2000 terms.}. 
We also conduct a user study concerning the explainability of learned logical rules. 
In our study, annotators agree that 
79\% (on average over three annotators) of the matched logical rules can be used to explain why a span is predicted as a target entity.

In summary, our main contributions are:
\begin{itemize}[nosep, leftmargin=*]
    \item We define five types of logical rules and introduce compound logical rules that can accurately detect entity boundaries and classify their types. Automatically learned rules can significantly reduce manual effort and provide explanations for entity predictions.

    \item To effectively learn rules, we propose a novel weakly supervised method with a dynamic label selection strategy that can 
    ensure the quality of pseudo labels.  
    
    \item We conduct experiments on both general and domain-specific datasets and  demonstrate the effectiveness of our method. 
    
\end{itemize}

\begin{figure*}[t]
\centering
 \includegraphics[width=\textwidth]{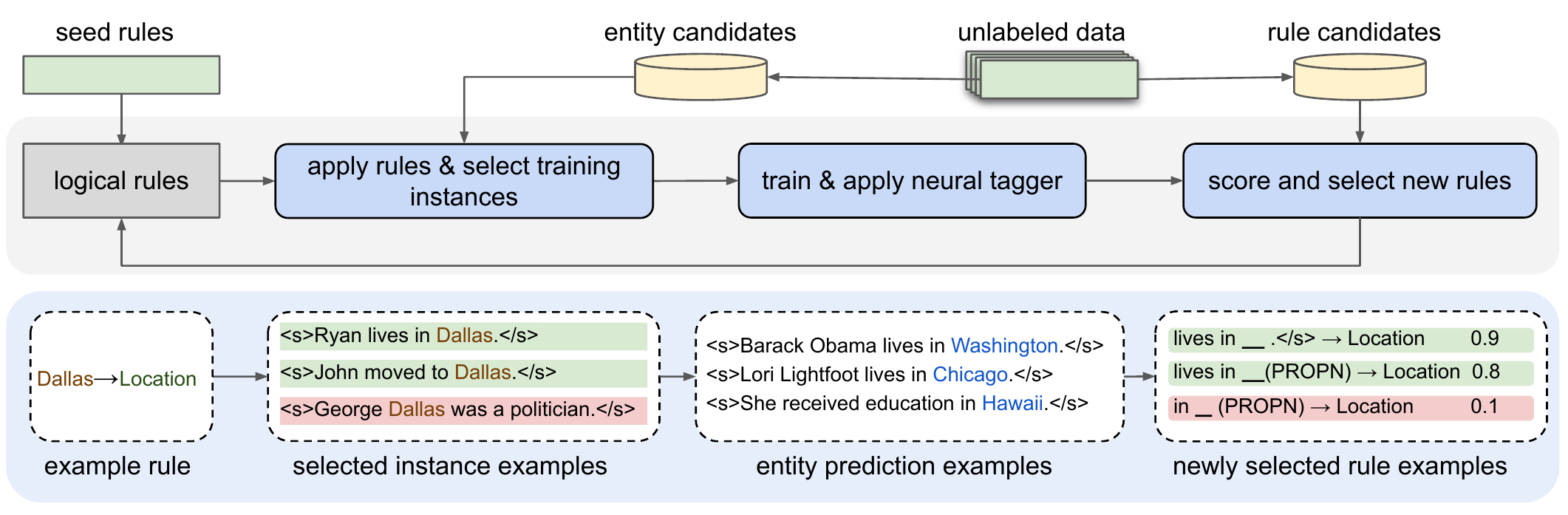}
\caption{
Overview of our tagging framework with logical rules and examples for each step. 
}
\label{fig:overview}
\end{figure*}

    
\section{Tagging with Learned Logical Rules}
We study named entity tagging under a weakly supervised setting, and propose
\textsc{TaLLoR} ( {\bf Ta}gging with {\bf L}earnable {\bf Lo}gical {\bf R}ules)
 to build a tagger with only a small set of rules. Compared with previous work, our framework requires less human effort 
via the use of
learned rules; 
we also show that these rules can be used to
explain tagging results.
Instead of treating tagging as a sequence labeling task, we formulate tagging as a span labeling task, in which named entities are modeled as spans over one or more tokens. With this setting, logical rules can easily be used for labeling entities.


\paragraph{Overview}
Figure~\ref{fig:overview} shows the flow of our iterative learning framework, which consists of 
the
following components. 
First, we generate all entity candidates and rule candidates from unlabeled data.
Then, for each iteration, 
we apply logical rules to the unlabeled data and select a set of high-quality weak training examples. 
Next, we train a neural tagger with the selected training examples and predict the labels of unlabeled data using the trained model. 
Finally, we select new accurate logical rules from candidate rules using the predictions. The newly learned rules will further be used to obtain weak training labels for the next iteration. 

\subsection{Logical Rule Extraction}
\label{rule}


In our work, a logical rule is defined in the form of 
``{\it if $p$ then $q$}'' (or ``$p \rightarrow q$'').\footnote{``heuristic rules'' and ``labeling rules'' can also be converted to logical rules, so they can be used interchangeably.}
For entity tagging, $q$ is one of the target entity classes, and $p$ can be any matching logic.
For example, ``{if a span's preceding tokens are `{\it lives in}', then it is a Location}''.
We design the following five 
types
of simple logical rules to consider the lexical, local context, and syntax information of an entity candidate.

\smallsection{Simple Logical Rules}
A simple logical rule is defined as a logical rule that contains a single condition predicate. We design the following five predicates to represent common logical conditions.
Given a candidate entity,
(1) \mtt{TokenString} matches its lexical string;
(2) \mtt{PreNgram} matches its preceding context tokens;
(3) \mtt{PostNgram} matches its succeeding context tokens; 
(4) \mtt{POSTag} matches its part-of-speech tags; 
(5) \mtt{DependencyRel} matches the dependency relations of its head word. 

\includegraphics[width=0.85\linewidth]{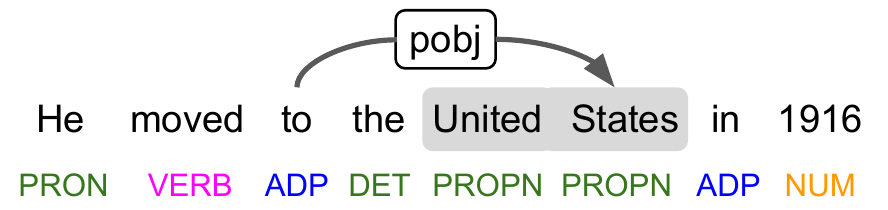}
Given a candidate entity ``{\it United States}'' in the above example, 
we can extract the following example logical rules for recognizing Locations:\footnote{All words in rules are lower-case and lemmatized.}
\begin{table}[h]
    \center
    \begin{tabular}{ll}
    \mtt{TokenString}==``\textit{united state}'' &$\rightarrow$ Location, \\
    \mtt{PreNgram}==``\textit{move to the}'' &$\rightarrow$ Location, \\
    \mtt{PostNgram}==``\textit{in 1916}'' &$\rightarrow$ Location, \\
    \mtt{POSTag}==``\textsc{propn propn}'' &$\rightarrow$ Location, \\
    \mtt{DependencyRel}==``\textit{to}'' (via pobj)&$\rightarrow$ Location. 
    \end{tabular}
\end{table}
%

More details about extraction of each condition predicate are included in Appendix~\ref{app:rules}.

%

\smallsection{Compound Logical Rules}
A compound logical rule is formed with multiple condition predicates and logical connectives including \mtt{and} ($\wedge$), \mtt{or} ($\vee$), and \mtt{negation} ($\neg$). In this work, we focus on learning compound logical rules connected with conjunctions ($\wedge$) to recognize entities precisely, because simple logical rules are often insufficient to identify entity boundaries. 
In the above example, the rule 
\mtt{PreNgram}==``\textit{move to the}'' 
can match multiple candidates such as 
``{\it United}'', ``{\it United States}'', and ``{\it United States in}'' etc., 
of which
many are inaccurate. 
However, with a compound rule, e.g.~\mtt{PreNgram}==``\textit{move to the}'' $\wedge$ 
\mtt{POSTag}==``\textsc{propn propn}'', 
we can correctly recognize that  ``{\it Unitied States}'' is a Location.

We enumerate and extract all possible logical rules from unlabeled data based on our pre-defined rule types before the training process.





\subsection{Applying Logical Rules}
At each iteration, we apply both seed and learned logical rules to unlabeled entity candidates to obtain a set of weakly labeled instances. 
In case an entity candidate is matched by multiple rules (potentially conflicting), we use the majority vote as the final weak label. 

\smallsection{Entity Candidates} 
In this work, we treat tagging as a span labeling task as described earlier. 
Before our learning process, we enumerate all token spans up to a maximum length from unlabeled data as entity candidates. 
%
We also notice that common phrases (e.g., ``{\it United States}'') are rarely split into different entities (e.g. ``{\it United}'', ``{\it States}''). Therefore, we generate a list of common phrases using the unsupervised AutoPhrase method \cite{Shang2018AutomatedPM} and merge two continuous spans together as a single entity candidate if they can form a common phrase.

\subsection{Dynamic Training Label Selection}
\label{eval module}

After applying the learned rules to unlabeled data, some of the weakly generated labels can be incorrect, which will lead to poor performance of our neural tagger in the next step.
To filter out noisy labels, we propose to maintain a high-precision entity set to keep the accurately labeled training examples from each iteration.

Inspired by \citet{Zhang2020EmpowerES}, we design a method to select high-quality labels from weakly generated labels by seed logical rules into the high-precision set.
Specifically, given an entity category $i$, its corresponding high-precision set $H_i$, and a weakly labeled instance $e_q$, we first compute a confidence score of $e_q$ belonging to category $i$ by considering both its maximum pair similarity to the high-precision set $H_i$ (called {\bf local score}) and its average similarity to $H_i$ (called {\bf global score}).
Then, the weakly labeled instance $e_q$ will be selected into the high-precision set if its confidence score 
is
larger than a threshold that is also estimated based on the high-precision set.

\smallsection{Instance Embedding}
We compute the embedding of an entity instance as the mean of the embeddings of its tokens. 
A
token's embedding is computed as the average of the first three layers' outputs from a pre-trained language model~\footnote{We used different pre-trained language models for different domains. Details are in Section~\ref{exp:setting}.}.

\smallsection{Local Score} 
Given a weakly labeled instance $e_q$ and an example $e_i$ from the high-precision set, we first compute their similarity as the cosine score between their embeddings.
Then, we compute the local confidence score of $e_q$ belonging to category $i$ as the maximum of its similarities between all examples in the high-precision set.


\smallsection{Global Score}
The local score is estimated based on a single instance in the high-precision set. Though it can help explore new entities, it can also be inaccurate in some cases.
Therefore, we propose to compute a more reliable score to estimate the accuracy of an instance $e_q$ belonging to a category $i$, which is called 
the
global score.
Specifically, we first sample a small set $E_s$ from the high precision set $H_i$ and then compute the prototypical embedding $\mathbf{x}_{E_s}$ of $E_s$ as the average of embeddings of all instances in $E_s$.
In our work, we sample $N$ times and compute the global score as:
\begin{align}
 \mathrm{score}^{\mathit{glb}}_{i}&= \frac{1}{N}\sum_{1\leq j\leq N} \mathbf{\cos}(\mathbf{x}^{j}_{E_s}, \mathbf{x}_{e_q})
\end{align}
To balance the exploration ability and reliability, we compute the final confidence score of a weakly labeled instance belonging to a category as the geometric mean of its local and global scores.

\smallsection{Dynamic Threshold Estimation}
We hypothesize that different categories of entities may have different thresholds for selecting high-quality weak labels. 
We may also need to use different thresholds at different iterations to dynamically balance exploration and reliability.
For example, we may expect our learning process to be reliable at earlier iterations and be exploratory at 
later stages. 
Motivated by this hypothesis, 
we propose to use a dynamic threshold to select high-quality weak labels. 
Specifically, we hold out one entity instance in the high precision set and compute its confidence score with respect to the rest of 
the
examples in the high-precision set. We randomly repeat $T$ times and use the minimum value as the threshold. For category $i$, it is calculated as:
\begin{equation}
    \mathrm{threshold} = \tau \cdot \min_{k\leq T, e_k \in H_i} \mathrm{score}_i (e_k)
\end{equation}
where $e_k$ is the held-out entity instance and $\tau \in [0, 1]$ is a temperature to control the final threshold. 

\subsection{Neural Tagging Model}

Following \citet{Jiang2020GeneralizingNL}, we treat tagging as a span labeling problem.
The key idea is to represent each span as a fixed-length embedding and make predictions based on its embedding. Briefly, given a span and its corresponding sentence, we first initialize all tokens in a sentence using a pre-trained language model, and then apply a Bi-LSTM and Self-Attention layer, and obtain the contextual embedding of the sentence. 
Finally, we compute the span embedding by concatenating two components: a {\it content representation} calculated as the weighted average across all token embeddings in the span, and a {\it boundary representation} that concatenates the embeddings at the start and end positions of the span.
Then, we predict the label of a span using a multilayer perceptron (MLP). For our detailed formulation please refer to Appendix~\ref{app:ner_model_detail}.

\subsection{Logical Rule Scoring and Selection}
Every iteration, we first predict the labels of all text spans using our neural tagging model. 
Then, we rank and select the $70\%$\footnote{
Different categories and datasets may require different thresholds to select high-quality labels. Setting a percentage means we will have dynamic thresholds for different categories so that the model will be robust to different categories and domains.}
most confident spans per category based on their prediction probabilities from the
tagging model as weak labels for computing rule scores.
We select new rules from rule candidates based on their confidence scores.
We adopt the {\it RlogF} method~\cite{Thelen2002ABM} to compute the confidence score of a rule $r$: \\
\begin{equation}
    F(r) = \frac{F_i}{N_i} \log_2(F_i) 
\end{equation}
where $F_i$ is the number of 
spans predicted with category label $i$ and matched by rule $r$,
and $N_i$ is the total number of spans matched by rule $r$. Intuitively, this method considers both the accuracy and coverage of rules because $\frac{F_i}{N_i}$ is the accuracy of the rule and $\log_2(F_i)$ represents the rule's ability to cover more spans.

In our experiments, we select 
the
top $\mathrm{K}$ rules for each entity class per iteration. 
We increase $\mathrm{K}$ by $\eta$ per iteration to be more exploratory in later iterations.  
We also use a threshold $\theta$ of rule accuracy (i.e.~$\frac{F_i}{N_i}$) to filter out noisy rules.
This method allows a variety of logical rules to be considered, yet is precise enough that all logical rules are strongly associated with the category.

    
\section{Experiments}
    \begin{table*}[t]
    \center
    \small
    \begin{tabular}{r|ccc|ccc|ccc}
        \toprule
         \multicolumn{1}{c|}{\multirow{2}{*}{Methods}}  & \multicolumn{3}{c}{BC5CDR}  & \multicolumn{3}{c}{CHEMDNER} & \multicolumn{3}{c}{CONLL2003}\\
                    
        \cmidrule{2-10}
             &  Precision & Recall & F$_1$ & Precision & Recall & F$_1$ &  Precision & Recall & F$_1$ \\
        \midrule
        Seed Rules & {\bf 94.09} & 3.81 & 7.33 & {\bf 91.60} & 13.19 & 23.07 & {\bf 95.77} & 2.76 & 5.36\\
        
        LinkedHMM & 10.18 & 15.60 & 12.32 & 23.99 & 10.77 & 14.86 & 19.78 & 31.51 & 24.30\\
        
        HMM-agg. & 43.70 & 21.60 & 29.00 & 49.60 & 18.40 & 26.80 & 52.00 & 8.50 & 14.60\\
        
        CGExpan & 40.96 & 24.75 & 30.86 & 45.70 & 25.58 & 32.80 & 55.97 & 28.7 & 37.95\\
        
        AutoNER & 42.22 & 30.66 & 35.52 & 66.83 & 27.59 & 39.05 & 32.07 & 5.98 & 10.08 \\ 
        
        Seed Rules + Neural Tagger & 78.33 & 21.60 & 33.86 & 84.18 & 21.91 & 34.78 & 72.57 & 24.68 & 36.83\\
        
        Self-training & 73.69 & 29.55 & 42.19 & 85.06 & 20.03 & 32.42 & 72.80 & 24.83 & 37.03\\  
        
        \midrule
    Our Learned Rules & 79.29 & 18.46 & 29.94 & 69.86 & 21.97 & 33.43 & 65.51 & 21.12 & 31.94\\
        Ours w/o Autophrase   & 74.56 & 32.93 & 45.68 & 67.74 & 55.99 & 61.31 & 71.37 & 25.50 & 37.57\\  
        Ours w/o Instance Selection & 58.70 & 63.37 & 60.95 & 42.64 & 48.25 & 45.27 & 58.51 & 58.8 & 58.65\\
        \our & 66.53 & {\bf 66.94} & {\bf 66.73} & 63.01 & {\bf 60.18} & {\bf 61.56} & 64.29 & {\bf 64.14} & {\bf 64.22}\\  
        \bottomrule
    \end{tabular}
    \caption{
   Performance of baselines (in upper section), our method and its sub-components (in lower section).
  }
    \label{tab:main_perf}
    \end{table*}
    
We first compare our method with baselines on three datasets and further analyze the importance of each component in an ablation study. 
We also report the performance of our method with different numbers of seed rules and at different iterations. 
Finally, we show an error analysis and present a user study to analyze how many logical rules can be used as understandable explanations.

\subsection{Experimental Setting}
\label{exp:setting}
We evaluate our method on the following three datasets. Note that we use each training set {\bf without} labels as our unlabeled data. 

 \noindent \textbf{BC5CDR}~\cite{Li2016BioCreativeVC} is the BioCreative V CDR task corpus. It contains 500 train, 500 dev, and 500 test PubMed articles, with 15,953 chemical and 13,318 disease entities.
\\ \noindent \textbf{CHEMDNER}~\cite{Krallinger2015TheCC} 
contains
10,000 PubMed abstracts with 84,355 chemical entities, in which the training/dev/test set contain 
 14,522/14,572/12,434 sentences respectively. 
\\ \noindent \textbf{CoNLL2003}~\cite{Sang2003IntroductionTT} consists of 14,041/3,250/3,453 sentences in the training/dev/test set extracted from Reuters news articles. 
We use Person, Location, and Organization entities in our experiments.\footnote{
We do not evaluate on Misc category because it does not represent a single semantic category, which cannot be represented with a small set of seed rules. 
}

\smallsection{Seed Rules and Parameters}
In our experiments, we set the maximum length of spans to 5,
and select the top $\mathrm{K}=20$ rules in the first iteration for BC5CDR and CoNLL2003, and $\mathrm{K}=60$ for the CHEMDNER dataset. 
Since it is relatively easy for users to manually give some highly accurate \mtt{TokenString} rules (i.e.,~entity examples), 
we use \mtt{TokenString} as seed rules for all experiments. 
To be specific, we manually select $20$ highly frequent \mtt{TokenString} rules as seeds for BC5CDR and CoNLL2003 and $40$ for CHEMDNER because of its large number of entities.
The manual seeds for each dataset are shown in Appendix~\ref{app:seeds}. 
For pre-trained language models, we use BERT~\cite{Devlin2019BERTPO} for CoNLL2003, and SciBert~\cite{Beltagy2019SciBERTAP} for BC5CDR and CHEMDNER. 
All our hyperparameters are selected on dev sets.
More setting details are in Appendix~\ref{app:parameters}.

\subsection{Compared Baseline Methods}

\noindent\textbf{Seed Rules}. We apply only seed rules to each test set directly and evaluate their performance.

\noindent\textbf{CGExpan}~\cite{Zhang2020EmpowerES} is a state-of-the-art lexicon expansion method by probing a language model.
Since \mtt{TokenString} seed rules can be viewed as a seed lexicon, 
we expand its size to 1,000 using this method and use them as \mtt{TokenString} rules.
We apply the top 200, 500, 800, and 1,000 rules to test sets and report the best performance. 

\noindent\textbf{AutoNER}~\cite{Shang2018LearningNE} takes lexicons of typed terms and untyped mined phrases as input. We use the best expanded lexicon from CGExpan as typed terms, 
and both of the expanded lexicon and the mined phrases from
AutoPhrase~\cite{Shang2018AutomatedPM} as untyped mined phrases. 
For detailed information on the AutoNER dictionary,
refer to Appendix%
~\ref{app:dict_autoner}

\noindent\textbf{LinkedHMM}~\cite{Safranchik2020WeaklySS} introduces a new generative model to incorporate noisy rules as supervision and predict entities using a neural NER model.
In our experiments, we use the expanded lexicon by CGExpan as tagging rules and AutoPhrase mined phrases as linking rules. 

\noindent\textbf{HMM-agg.}~\cite{Lison2020NamedER}
proposes a hidden Markov model to first generate weak labels from labeling functions and train a sequence tagging model. We convert the expanded lexicon by CGExpan to labeling functions and report results of the tagging model.


\noindent\textbf{Seed Rule + Neural Tagger}. This method is our framework without iteration learning. 
After applying seed rules, we use the weakly generated labels to train our neural tagger and report the result of the tagger.

\noindent\textbf{Self-training}. 
We first obtain weak labels by applying seed rules. Then, we build a 
self-training system  using the weak labels as initial supervision and our neural tagger as the base model. 

Methods~\cite{Fries2017SwellSharkAG, Ratner2017SnorkelRT, Huang2010InducingDS} which use noun phrases as entity candidates are not included here because noun phrases have poor recall on the three datasets as shown in Table~\ref{tab:bound_detection}. CGExpan outperforms other entity set expansion methods (e.g.,~\citet{Yan2019LearningTB}) so we use CGExpan as our baseline for automatic lexicon expansion.




\begin{figure*}[t]
     \centering
     \begin{subfigure}{0.325\textwidth}
         \centering
         \includegraphics[width=\textwidth]{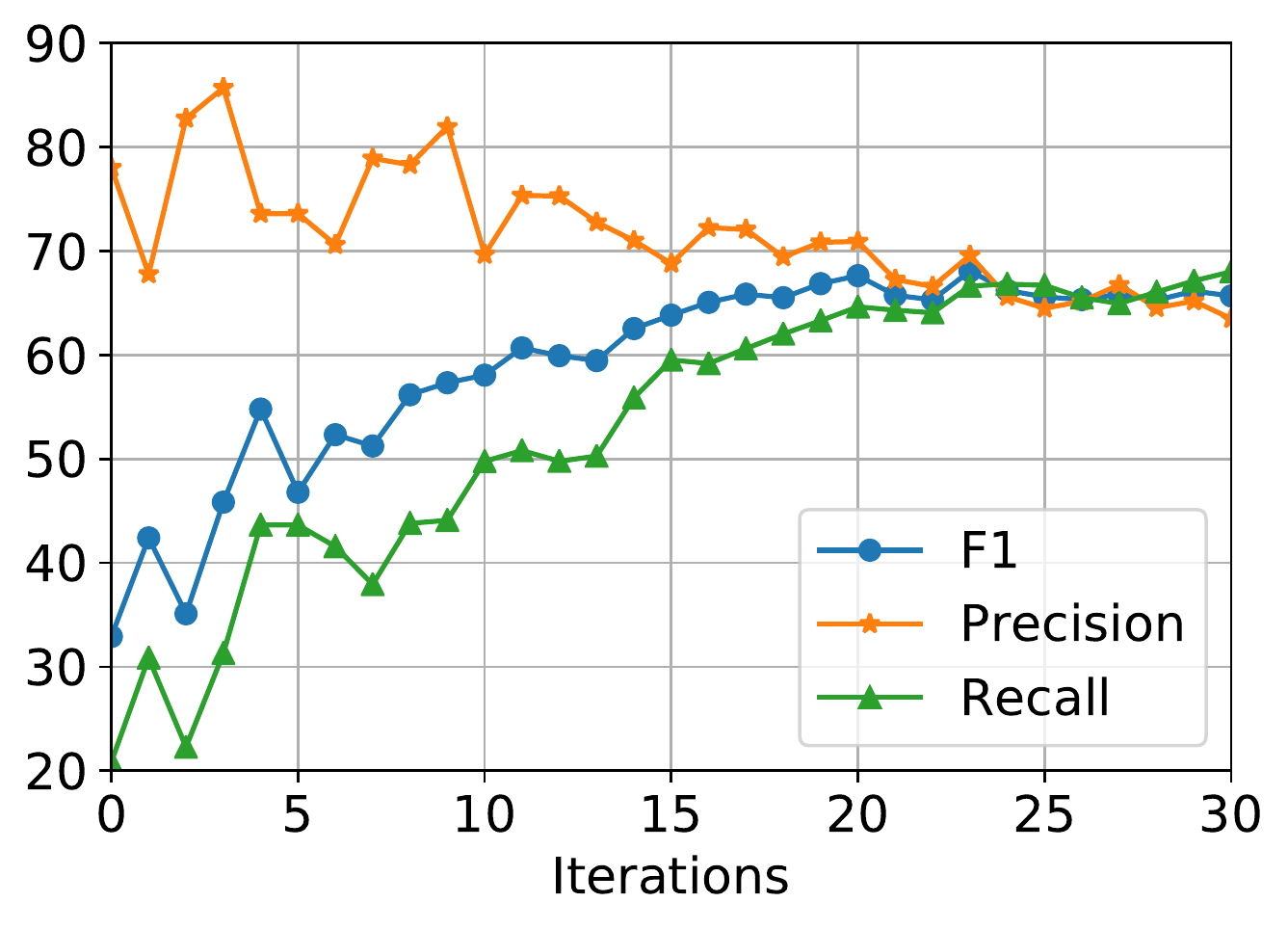}
         \caption{}
         \label{fig:bc5cdr_iter_f1}
     \end{subfigure}
     \hfill
     \begin{subfigure}{0.325\textwidth}
         \centering
         \includegraphics[width=\textwidth]{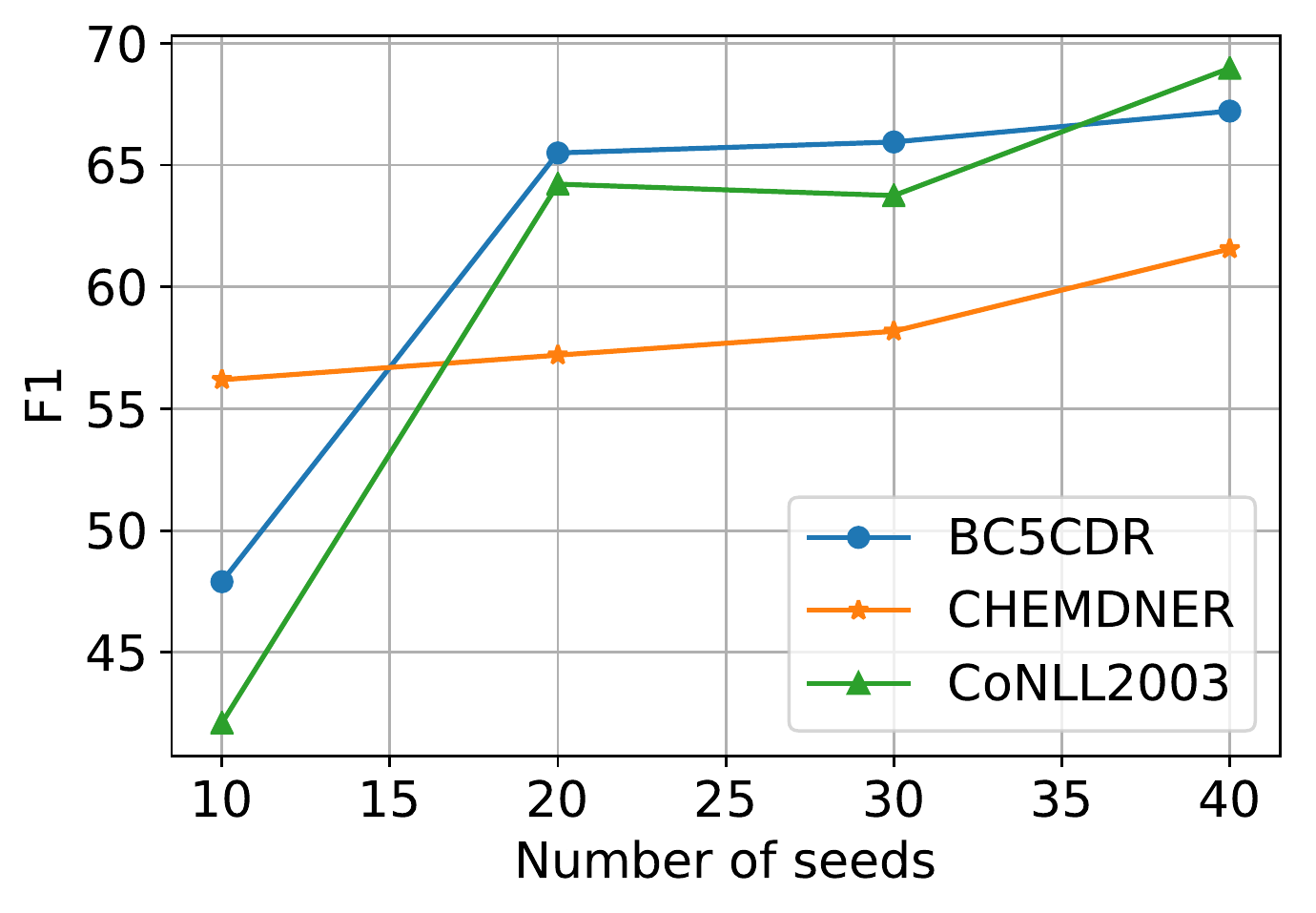}
         \caption{}
         \label{fig:diff_seed}
     \end{subfigure}
     \hfill
     \begin{subfigure}{0.325\textwidth}
         \centering
         \includegraphics[width=\textwidth]{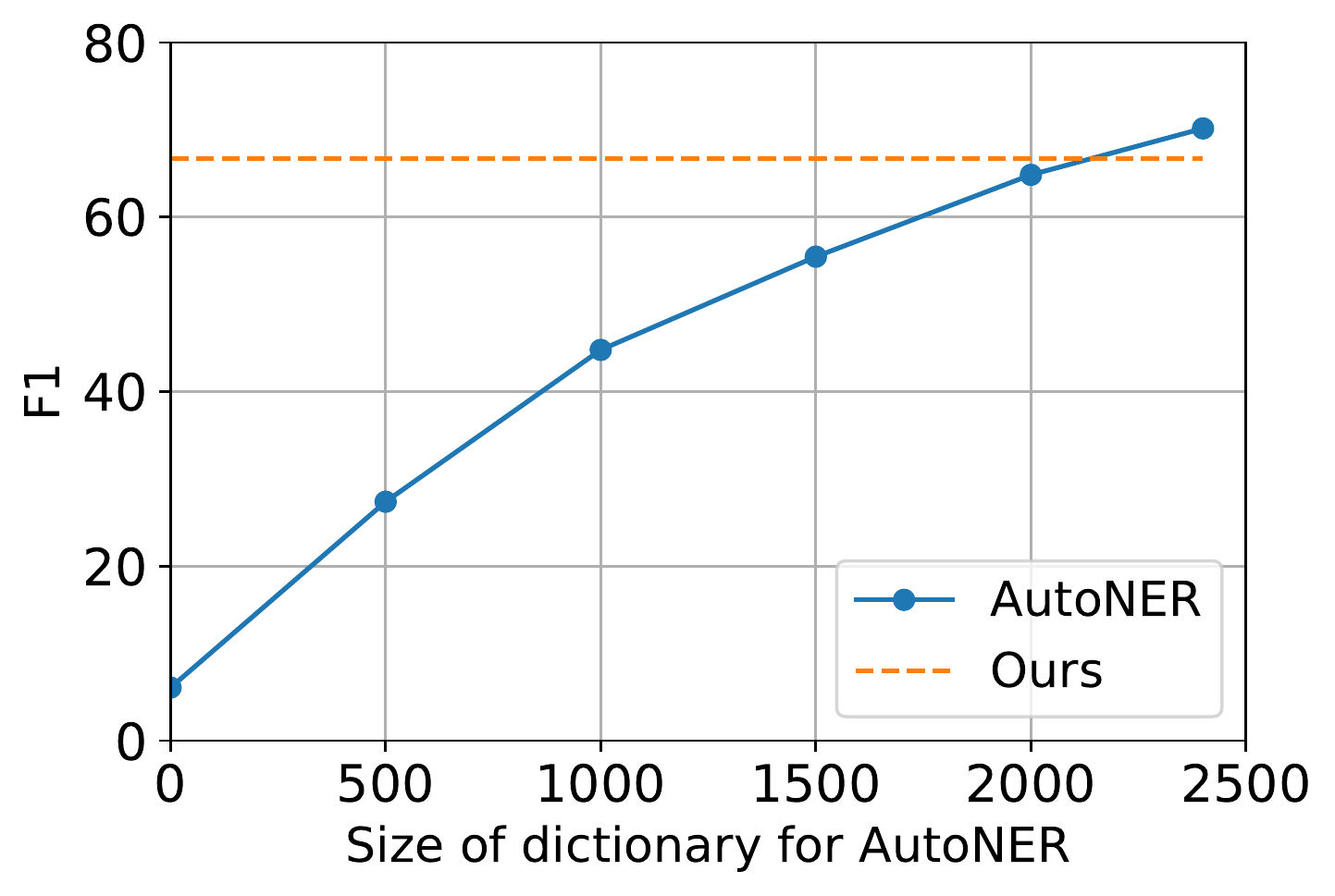}
         \caption{}
         \label{fig:compare_distant}
     \end{subfigure}
    \caption{(a) Iterations vs.~performance of our method on BC5CDR. (b) Performance with different numbers of seed rules.  (c) Performance of AutoNER with different sizes of manual lexicon and our method on BC5CDR.
    }
\end{figure*}

\subsection{Performance of Compared Methods}
We present the precision, recall, and micro-averaged $F_1$ scores on three datasets in Table~\ref{tab:main_perf}. 
Results show that our method significantly outperforms baseline methods obtaining an average of 
$24$-point
$F_1$ improvement across three datasets over the best baseline. 

We see that the precision of our seed rules is high, but the recall is lower.
The lexicon expansion method (CGExpan) can recognize more entities but also introduces errors resulting in an improvement to recall but a dramatic decrease in precision. 


Existing weakly supervised methods (i.e.,~AutoNER, LinkedHMM and HMM-agg.) cannot recognize entities effectively with either seed rules or expanded rules by CGExpan.
These methods require
a
high-precision lexicon as input;
however, the precision of the automatically expanded lexicon is not 
sufficient
to meet this requirement. 
Though seed rules are very accurate, they lack 
coverage of various entities. 

Our method without iteration ({\it Seed Rules + Neural Tagger}) and self-training can achieve high precision because 
of the accurate pseudo labels generated from seed rules.
It is interesting to note that the self-training method based on our neural tagger also achieved low recall. We hypothesize that this is mainly due to 
the neural tagger 
overfitting the small set of 
labels from seed rules. 


\smallsection{Ablation Study}
We also performed an ablation study to analyze the importance of some components in our framework, and report the performance in  
Table~\ref{tab:main_perf} (the lower section).
Results show that our learned rules are accurate but lack 
coverage.
Without using 
common phrases mined by Autophrase (i.e., {\it Ours w/o Autophrase}), our method achieves dramatically lower recall demonstrating the effectiveness of common phrases for improving 
coverage. 
Without high-quality training instance selection ({\it Ours w/o Instance Selection}), the precision is lower than our best method indicating the importance of
the
instance selection step.


\subsection{Analysis of Learning Iterations and Seeds}
\smallsection{Performance vs.~Iterations}
Figure~\ref{fig:bc5cdr_iter_f1} shows the performance of our method at different iterations. We see that our method improves the recall from $20\%$ to over $60\%$ during the learning process with 
a slight
decease in precision,
and achieves the best $F_1$ score after 25 iterations. 
Results on other two datasets show the same trend (in Appendix~\ref{app:perf_iter}).


\smallsection{Performance with Different Numbers of Seeds}
\label{diff seed}
Figure~\ref{fig:diff_seed} shows the performance of our method using different numbers of manually selected seed rules on three datasets. 
We see that our method can achieve continuous improvement using more seeds. 
We also notice that our method can achieve over $55\%$ $F_1$ on CHEMDNER with only 10 seeds demonstrating the effectiveness of our framework under minimal supervision setting. 
Our method obtains significantly better results (around $65\%$ $F_1$) when using 20 seeds than 10 seeds on BC5CDR and CoNLL indicating that 20 seeds is a reasonable starting point for building a tagging system without much manual effort. 

\subsection{Comparison with Distant Supervision}
AutoNER \cite{Shang2018LearningNE} is a distantly supervised method using a manually created lexicon as supervision. 
We also compared our method to this method to figure out how many terms we need to manually created for AutoNER to achieve similar performance with our method. 
We conducted experiments on BC5CDR and used only 20 seeds for our method. 
For AutoNER, we used additional $\mathrm{M}$ terms from a manually created lexicon \cite{Shang2018LearningNE}\footnote{AutoNER authors compiled the lexicon from MeSH database and CTD Chemical and Disease vocabularies, which are manually created by experts.}.
Figure~\ref{fig:compare_distant} shows the performance with different values of $\mathrm{M}$. 
Results show that AutoNER needs an additional 
$\sim{}2000$ terms to achieve similar performance (around $66\%$ $F_1$) with our method, which demonstrates that our method is effective under minimal supervision without access to a large 
manual lexicon.

\begin{table*}[h]
    \centering
    \small
    \begin{tabular}{p{10cm}|c|c}
        \toprule
      {\bf Examples} & {\bf Predicted Labels} & {\bf Gold Label} \\ 
      \midrule
   \multicolumn{3}{c}{ {\bf Error Type:  Similar Semantic Concepts (56\%)} }     \\ 
         The aim of this work is to call attention to the risk of {\color{red}tacrolimus} use in patients with SSc.
         &  Disease  & NotEntity \\
         We recorded time to first dysrhythmia occurrence , respective times to 25 \% and 50 \% reduction of the {\color{red}heart rate} ( HR ) and mean arterial pressure , and time to asystole and total amount of bupivacaine consumption. 
        &  Disease  & NotEntity \\
        The severity of pain due to etomidate injection , mean arterial pressure , heart rate , and {\color{red}adverse effects} were also evaluated. 
        & Disease  & NotEntity\\
         \midrule
   \multicolumn{3}{c}{ {\bf Error Type:  Inaccurate Boundary (20\%)  } }     \\
        Furthermore ameliorating effect of crocin on diazinon induced {\color{red}{disturbed}} \underline{{\color{red}cholesterol}} homeostasis was studied. 
        &  Disease  & Disease \\
        Pretreatment with S. virgaurea extract for 5 weeks at a dose of 250 mg / kg followed by \underline{{\color{red}isoproterenol}} {\color{red}injection} significantly prevented the observed alterations. &  Chemical  & Chemical\\
        This depressive -like profile induced {\color{red} by} \underline{{\color{red}METH}} was accompanied by a marked depletion of frontostriatal dopaminergic and serotonergic neurotransmission , indicated by a reduction in the levels of dopamine , DOPAC and HVA , tyrosine hydroxylase and serotonin , observed at both 3 and 49 days post - administration. 
         & Chemical  &  Chemical\\
         \midrule
   \multicolumn{3}{c}{ {\bf Error Type:  Nested Entity (20\%)  } }     \\  
        Early \underline{postoperative {\color{red}delirium}} incidence risk factors were then assessed through three different multiple regression models.
        & Disease & Disease \\
        The impact of immune - mediated heparin -induced thrombocytopenia type II (\underline{{\color{red}HIT} type II} ) as a cause of thrombocytopenia.
        & Disease & Disease \\
        Extensive literature search revealed multiple cases of \underline{coronary artery {\color{red}vasospasm}} secondary to zolmitriptan , but none of the cases were associated with TS.
         & Disease   &  Disease\\
         \midrule
   \multicolumn{3}{c}{ {\bf Error Type:  Others (4\%) } }     \\
        It is characterized by its intense urotoxic action , leading to {\color{red}hemorrhagic cystitis}.
         & Disease  & NotEntity\\
         Famotidine is a {\color{red} histamine} H2-receptor antagonist used in inpatient settings for prevention of stress ulcers and is showing increasing popularity because of its low cost .
         & Chemical  & NotEntity\\
         It is characterized by its intense urotoxic action , leading to {\color{red} hemorrhagic cystitis}.
         & Disease  & NotEntity\\
         \bottomrule
    \end{tabular}
    \caption{
    Gold entities are underlined, predicted entities are in red. 
    {\bf Error type ``similar semantic concepts''} means that our rules cannot distinguish two closely related semantic concepts. 
    {\bf Error type ``inaccurate boundary''} means our rules label incorrectly about the boundaries of entities. 
    {\bf Error type ``nested entity''} means the error is due to multiple possible entities are nested. 
    {\bf NotEntity} means the predicted span is not an entity. 
    }
     \label{tab:error_example}
\end{table*}
\subsection{Analysis of Rule Selection Strategies}

\begin{table}[h]
    \centering
    \small
    \begin{tabular}{r|ccc}
        \toprule
       Rule Selection Strategy & Precision & Recall & $F_1$\\      
        \midrule
         Rule Type & 57.14 & 63.00 & 59.93\\
         Entity\&Rule Type & 61.73 & 64.97 & 63.31\\
         Entity Type & \textbf{66.53} & \textbf{66.94} & \textbf{66.73}\\
        \bottomrule
    \end{tabular}
    \caption{Performance on the BC5CDR dataset with three different rule selection strategies.
    }
    \label{tab:strategies}
    \vspace{-1mm}
\end{table}

In our work, we designed three rule selection strategies:
(1) {\it entity type}
selects the top $\mathrm{K}$ rules for each entity category; 
(2) {\it rule type} 
selects the top $\mathrm{K}$ rules for each logical rule type; 
(3) {\it entity\&rule type}
selects the top $\mathrm{K}$ rules for each entity category and logical rule type.
Results in Table~\ref{tab:strategies} show that entity type based selection achieves the best performance. 


\begin{table}[t]
    \centering
    \small
    \begin{tabular}{r|ccc}
        \toprule
        Rule Type &  BC5CDR &  CHEMDNER &  CoNLL \\
        \midrule
         TokenStr & 503 (41\%) & 1667 (44\%) & 779 (25\%)\\
         Pre $\wedge$ Post & 203 (17\%) & 629 (17\%) & 956 (31\%)\\
         Pre $\wedge$ POS & 288 (24\%) & 585 (16\%) & 455 (15\%)\\
         POS $\wedge$ Post & 149 (12\%) & 418 (11\%) & 438 (14\%) \\
         Dep $\wedge$ POS & 79 (6\%) & 432 (12\%) & 469 (15\%) \\
        \bottomrule
    \end{tabular}
    \caption{Number and ratio of different type rules.}
    \label{tab:num_rules}
    \vspace{-2mm}
\end{table}

%
%

\begin{table*}[t]
    \center
    \small
    \begin{tabular}{llc}
        \toprule
      Labeled Entities and Sentences & Learned Logical Rules & Entity type \\
        \midrule
        \makecell[l]{This occlusion occurred after EACA therapy in {\color{applegreen}a patient with}\\{\color{blue}SAH} {\color{applegreen}and} histopathological documentation of recurrent SAH.}
        & \makecell[l]{\mtt{PreNgram}=``a patient with'' \\
        ~~~~~~$\land$ \mtt{PostNgram}=``and''} 
        & \multirow{2}{*}{Disease} \\
       \hline
        \makecell[l]{We also analyzed published and unpublished follow-up data to\\determine {\color{applegreen}the risk of} $\stackrel{\text{ {\color{applegreen} PROPN}} }{{\color{blue}\text{ICH}} }$ in antithrombotic users with MB.}
        & \makecell[l]{\mtt{PreNgram}=``the risk of'' \\ 
        ~~~~~~$\land$ \mtt{POStag}=\textsc{propn} }
        & \multirow{2}{*}{Disease} \\
        \hline
        \makecell[l]{3 weeks after initiation of amiodarone {\color{applegreen}therapy for} $\stackrel{\text{ {\color{applegreen} ADJ}} }{{\color{blue}\text{atrial}} }$ $\stackrel{\text{ {\color{applegreen} NOUN}} }{{\color{blue}\text{fibrillation}}}$.}
        & \makecell[l]{\mtt{PreNgram}=``therapy for'' \\
        ~~~~~~$\land$ \mtt{POStag}=\textsc{adj noun}}
        & \multirow{2}{*}{Disease}\\
        \hline
        \makecell[l]{Although 25 {\color{applegreen}mg of} 
        $\stackrel{\text{ {\color{applegreen} NOUN}} }{{\color{blue}\text{lamivudine}} }$ was slightly less effective than\\ 100mg (P=.011) and 300 mg ( P=.005).}
        & \makecell[l]{ \mtt{PreNgram}=``mg of'' \\ 
        ~~~~~~$\land$ \mtt{POStag}=\textsc{noun}}
        & \multirow{2}{*}{Chemical}\\
        \hline
        \makecell[l]{These results suggest that the renal {\color{applegreen}protective effects of}
        $\stackrel{\text{ {\color{applegreen} NOUN}} }{{\color{blue}\text{misoprostol}} }$
        \\ is dose - dependent.}
        & \makecell[l]{\mtt{PreNgram}=``protective effect of'' \\
        ~~~~~~$\land$ \mtt{POStag}=\textsc{noun}}
        & \multirow{2}{*}{Chemical}\\
        \bottomrule
    \end{tabular}
    \caption{Examples of learned rules and correctly labeled entities (in red) by the learned rules in BC5CDR dataset.}
    \label{tab:rule_example}
\end{table*}

\subsection{Error Analysis of Learned Logical Rules}
We show the statistics of different types of rules learned after all iterations in Table~\ref{tab:num_rules}.\footnote{TokenStr, Pre, Post, POSTag, and Dep are short for TokenString, PreNgram, PostNgram, POSTag, and DependencyRel.} 
We see that TokenString rule is the most rule type for domain-specific datasets (BC5CDR and CHEMDNER). For the general domain task, PreNgram$\wedge$PostNgram is the most rule type learned by our model.

We also performed an error analysis on 
the
BC5CDR dataset. 
Specifically, we sampled $100$ entities predicted incorrectly by our learned rules and analyzed their error types.
Analysis results show that 56\% of errors are caused by an inability to distinguish closely related entity categories (chemicals vs medications), 
and another 20\% are due to incorrect detection of entity boundaries. 
We also notice that some spans (e.g.~``HIT type II'') and their sub-spans (e.g.~``HIT'') are both disease entities (i.e., nested entities), but only the longer ones are annotated with gold labels.
Our rules sometimes only predict the sub-spans as diseases, which contributes to 20\% of the errors. 
We put examples of each error type in Table~\ref{tab:error_example}.


\subsection{User Study of Explainable Logical Rules}

Since our logical rules are intuitive clues for recognizing entities, we hypothesize that automatically learned rules can be used as understandable explanations for the predictions of entities.
Therefore, we conducted a user study to find out how many logical rules are explainable. 
Specifically, we applied the learned rules in BC5CDR and 
sampled $100$ entities labeled by at least one logical rule other than 
TokenString\footnote{We exclude TokenString rules because they are self-explainable.} for our user study. Some examples are shown in Table~\ref{tab:rule_example}.
We asked two annotators
without domain knowledge and one biological expert to annotate whether our learned logical rules can be understood and used as explanations for why a span is predicted as a disease or chemical. 
Manual annotation results show that the two annotators
and the biological expert agree that 81\%, 87\%, and 70\% of the predicted entities can be explained by logical rules, respectively. 


\section{Related Work}

Different types of methods have been proposed to build named entity tagging systems using indirect or limited supervision. 
Distant supervision~\cite{Mintz2009DistantSF} is one kind of methods that have been proposed to alleviate human effort by training models using existing lexicons or knowledge bases.
Recently, there have been attempts to build NER systems with distant supervision~\cite{Ren2015ClusTypeEE, Fries2017SwellSharkAG, Giannakopoulos2017UnsupervisedAT}. 
AutoNER~\cite{Shang2018LearningNE} trained a NER system by using both typed lexicons and untyped mined phrases as supervision.
\citet{Peng2019DistantlySN} proposed an AdaPU algorithm to incorporate an incomplete dictionary as supervision. 
However, lexicons or knowledge bases are not always available for new domains and tasks, especially in specific domains and low-resource settings. Manually constructing these lexicons is often very expensive. 

Bootstrapping is a technique to learn models
from a small set of seeds, which has been proposed for 
word sense disambiguation~\cite{Yarowsky1995UnsupervisedWS} and product attribute extraction~\cite{putthividhya-hu-2011-bootstrapped}.
Bootstrapping methods
\cite{niu2003bootstrapping,Huang2010InducingDS}
have been proposed for building entity tagging systems by 
assuming target entities are just proper names or noun phrases. 
\citet{Gupta2014ImprovedPL} used an improved pattern scoring method to bootstrap domain-specific terminologies 
with restricted part-of-speech patterns. 
However,
previous works only focused on disambiguating entity types by assuming target entities are given or just syntactic chunks. 
But, as we shown earlier, target entities often do not align well with simple syntactic chunks. 
Bootstrapping methods that can automatically detect entity boundaries and predict their types simultaneously are desirable in real-world applications. 

Recently, methods have been proposed to obtain weak labels by manually writing labeling functions~\cite{Bach2017LearningTS}.
Based on this idea, several methods~\cite{Safranchik2020WeaklySS,Lison2020NamedER} have been proposed for NER by assuming the availability of a sufficient amount of handcrafted labeling functions and lexicons. 
However, manually designing labeling rules is challenging, which requires a significant amount of manual effort and domain expertise.
Our work aims to learn logical rules automatically to reduce human effort.

\section{Conclusion}

In this work, we explored how to build a tagger from a small set of seed logical rules and unlabeled data. 
We defined five types of simple logical rules and 
introduced compound logical rules that are composed from simple rules to detect entity boundaries and classify their types simultaneously. 
We also design a dynamic label selection method to select accurate pseudo labels generated from learned rules for training a discriminative tagging model.
%
%
Experimental results demonstrate that our method is effective and outperforms existing weakly supervised methods.




\newpage
\bibliography{anthology,acl2021}

\begin{thebibliography}{28}
\expandafter\ifx\csname natexlab\endcsname\relax\def\natexlab#1{#1}\fi

\bibitem[{Bach et~al.(2017)Bach, He, Ratner, and R{\'e}}]{Bach2017LearningTS}
Stephen~H. Bach, B.~He, A.~Ratner, and C.~R{\'e}. 2017.
\newblock Learning the structure of generative models without labeled data.
\newblock \emph{Proceedings of machine learning research}, 70:273--82.

\bibitem[{Beltagy et~al.(2019)Beltagy, Lo, and Cohan}]{Beltagy2019SciBERTAP}
Iz~Beltagy, Kyle Lo, and Arman Cohan. 2019.
\newblock Scibert: A pretrained language model for scientific text.
\newblock In \emph{EMNLP/IJCNLP}.

\bibitem[{Devlin et~al.(2019)Devlin, Chang, Lee, and
  Toutanova}]{Devlin2019BERTPO}
J.~Devlin, Ming-Wei Chang, Kenton Lee, and Kristina Toutanova. 2019.
\newblock Bert: Pre-training of deep bidirectional transformers for language
  understanding.
\newblock In \emph{NAACL-HLT}.

\bibitem[{Fries et~al.(2017)Fries, Wu, Ratner, and
  R{\'e}}]{Fries2017SwellSharkAG}
Jason~Alan Fries, Sen Wu, A.~Ratner, and Christopher R{\'e}. 2017.
\newblock Swellshark: A generative model for biomedical named entity
  recognition without labeled data.
\newblock \emph{ArXiv}, abs/1704.06360.

\bibitem[{Giannakopoulos et~al.(2017)Giannakopoulos, Musat, Hossmann, and
  Baeriswyl}]{Giannakopoulos2017UnsupervisedAT}
Athanasios Giannakopoulos, C.~Musat, Andreea Hossmann, and Michael Baeriswyl.
  2017.
\newblock Unsupervised aspect term extraction with b-lstm \& crf using
  automatically labelled datasets.
\newblock In \emph{WASSA@EMNLP}.

\bibitem[{Gupta and Manning(2014)}]{Gupta2014ImprovedPL}
S.~Gupta and Christopher~D. Manning. 2014.
\newblock Improved pattern learning for bootstrapped entity extraction.
\newblock In \emph{CoNLL}.

\bibitem[{Huang and Riloff(2010)}]{Huang2010InducingDS}
Ruihong Huang and E.~Riloff. 2010.
\newblock Inducing domain-specific semantic class taggers from (almost)
  nothing.
\newblock In \emph{ACL}.

\bibitem[{Jiang et~al.(2020)Jiang, Xu, Araki, and
  Neubig}]{Jiang2020GeneralizingNL}
Zhengbao Jiang, W.~Xu, J.~Araki, and Graham Neubig. 2020.
\newblock Generalizing natural language analysis through span-relation
  representations.
\newblock In \emph{ACL}.

\bibitem[{Krallinger et~al.(2015)Krallinger, Rabal, Leitner, Vazquez, Salgado,
  Lu, Leaman, Lu, Ji, Lowe, Sayle, Batista-Navarro, Rak, Huber,
  Rockt{\"a}schel, Matos, Campos, Tang, Xu, Munkhdalai, Ryu, Ramanan, Nathan,
  Zitnik, Bajec, Weber, Irmer, Akhondi, Kors, Xu, An, Sikdar, Ekbal, Yoshioka,
  Dieb, Choi, Verspoor, Khabsa, Giles, Liu, Ravikumar, Lamurias, Couto, Dai,
  Tsai, Ata, Can, Usie, Alves, Segura-Bedmar, Mart{\'i}nez, Oyarz{\'a}bal, and
  Valencia}]{Krallinger2015TheCC}
Martin Krallinger, O.~Rabal, Florian Leitner, Miguel Vazquez, David Salgado,
  Zhiyong Lu, Robert Leaman, Yanan Lu, Dong-Hong Ji, D.~M. Lowe, R.~Sayle,
  R.~Batista-Navarro, R.~Rak, Torsten Huber, Tim Rockt{\"a}schel, S{\'e}rgio
  Matos, D.~Campos, Buzhou Tang, H.~Xu, Tsendsuren Munkhdalai, K.~Ryu, S.~V.
  Ramanan, P.~S. Nathan, S.~Zitnik, M.~Bajec, L.~Weber, Matthias Irmer,
  S.~Akhondi, J.~Kors, S.~Xu, X.~An, Utpal~Kumar Sikdar, A.~Ekbal, M.~Yoshioka,
  Thaer~M. Dieb, Miji Choi, Karin~M. Verspoor, Madian Khabsa, C.~Lee Giles,
  H.~Liu, K.~Ravikumar, Andre Lamurias, F.~Couto, Hong-Jie Dai, R.~Tsai,
  C.~Ata, T.~Can, Anabel Usie, Rui Alves, Isabel Segura-Bedmar, Paloma
  Mart{\'i}nez, J.~Oyarz{\'a}bal, and A.~Valencia. 2015.
\newblock The chemdner corpus of chemicals and drugs and its annotation
  principles.
\newblock \emph{Journal of Cheminformatics}, 7:S2 -- S2.

\bibitem[{Lee et~al.(2017)Lee, He, Lewis, and Zettlemoyer}]{Lee2017EndtoendNC}
Kenton Lee, Luheng He, M.~Lewis, and Luke Zettlemoyer. 2017.
\newblock End-to-end neural coreference resolution.
\newblock In \emph{EMNLP}.

\bibitem[{Li et~al.(2016)Li, Sun, Johnson, Sciaky, Wei, Leaman, Davis,
  Mattingly, Wiegers, and Lu}]{Li2016BioCreativeVC}
J.~Li, Yueping Sun, Robin~J. Johnson, Daniela Sciaky, Chih-Hsuan Wei, Robert
  Leaman, A.~P. Davis, C.~Mattingly, Thomas~C. Wiegers, and Zhiyong Lu. 2016.
\newblock Biocreative v cdr task corpus: a resource for chemical disease
  relation extraction.
\newblock \emph{Database: The Journal of Biological Databases and Curation},
  2016.

\bibitem[{Lison et~al.(2020{\natexlab{a}})Lison, Hubin, Barnes, and
  Touileb}]{Lison2020NamedER}
P.~Lison, A.~Hubin, Jeremy Barnes, and Samia Touileb. 2020{\natexlab{a}}.
\newblock Named entity recognition without labelled data: A weak supervision
  approach.
\newblock \emph{ArXiv}, abs/2004.14723.

\bibitem[{Lison et~al.(2020{\natexlab{b}})Lison, Barnes, Hubin, and
  Touileb}]{lison-etal-2020-named}
Pierre Lison, Jeremy Barnes, Aliaksandr Hubin, and Samia Touileb.
  2020{\natexlab{b}}.
\newblock \href {https://doi.org/10.18653/v1/2020.acl-main.139} {Named entity
  recognition without labelled data: A weak supervision approach}.
\newblock In \emph{Proceedings of the 58th Annual Meeting of the Association
  for Computational Linguistics}, pages 1518--1533, Online. Association for
  Computational Linguistics.

\bibitem[{Mintz et~al.(2009)Mintz, Bills, Snow, and
  Jurafsky}]{Mintz2009DistantSF}
M.~Mintz, Steven Bills, R.~Snow, and Dan Jurafsky. 2009.
\newblock Distant supervision for relation extraction without labeled data.
\newblock In \emph{ACL/IJCNLP}.

\bibitem[{Niu et~al.(2003)Niu, Li, Ding, and Srihari}]{niu2003bootstrapping}
Cheng Niu, Wei Li, Jihong Ding, and Rohini~K Srihari. 2003.
\newblock A bootstrapping approach to named entity classification using
  successive learners.
\newblock In \emph{Proceedings of the 41st Annual Meeting of the Association
  for Computational Linguistics}, pages 335--342.

\bibitem[{Peng et~al.(2019)Peng, Xing, Zhang, Fu, and
  Huang}]{Peng2019DistantlySN}
Minlong Peng, Xiaoyu Xing, Qi~Zhang, Jinlan Fu, and X.~Huang. 2019.
\newblock Distantly supervised named entity recognition using
  positive-unlabeled learning.
\newblock \emph{ArXiv}, abs/1906.01378.

\bibitem[{Pennington et~al.(2014)Pennington, Socher, and
  Manning}]{Pennington2014GloveGV}
Jeffrey Pennington, R.~Socher, and Christopher~D. Manning. 2014.
\newblock Glove: Global vectors for word representation.
\newblock In \emph{EMNLP}.

\bibitem[{Putthividhya and Hu(2011)}]{putthividhya-hu-2011-bootstrapped}
Duangmanee Putthividhya and Junling Hu. 2011.
\newblock \href {https://www.aclweb.org/anthology/D11-1144} {Bootstrapped named
  entity recognition for product attribute extraction}.
\newblock In \emph{Proceedings of the 2011 Conference on Empirical Methods in
  Natural Language Processing}, pages 1557--1567, Edinburgh, Scotland, UK.
  Association for Computational Linguistics.

\bibitem[{Ratner et~al.(2017)Ratner, Bach, Ehrenberg, Fries, Wu, and
  R{\'e}}]{Ratner2017SnorkelRT}
A.~Ratner, Stephen~H. Bach, Henry~R. Ehrenberg, Jason~Alan Fries, Sen Wu, and
  C.~R{\'e}. 2017.
\newblock Snorkel: Rapid training data creation with weak supervision.
\newblock \emph{Proceedings of the VLDB Endowment. International Conference on
  Very Large Data Bases}, 11 3:269--282.

\bibitem[{Ren et~al.(2015)Ren, El-Kishky, Wang, Tao, Voss, and
  Han}]{Ren2015ClusTypeEE}
Xiang Ren, Ahmed El-Kishky, C.~Wang, Fangbo Tao, Clare~R. Voss, and Jiawei Han.
  2015.
\newblock Clustype: Effective entity recognition and typing by relation
  phrase-based clustering.
\newblock \emph{KDD : proceedings. International Conference on Knowledge
  Discovery \& Data Mining}, 2015:995--1004.

\bibitem[{Safranchik et~al.(2020)Safranchik, Luo, and
  Bach}]{Safranchik2020WeaklySS}
Esteban Safranchik, Shiying Luo, and Stephen~H. Bach. 2020.
\newblock Weakly supervised sequence tagging from noisy rules.
\newblock In \emph{AAAI}.

\bibitem[{Sang and Meulder(2003)}]{Sang2003IntroductionTT}
E.~T.~K. Sang and F.~D. Meulder. 2003.
\newblock Introduction to the conll-2003 shared task: Language-independent
  named entity recognition.
\newblock \emph{ArXiv}, cs.CL/0306050.

\bibitem[{Shang et~al.(2018{\natexlab{a}})Shang, Liu, Jiang, Ren, Voss, and
  Han}]{Shang2018AutomatedPM}
Jingbo Shang, Jialu Liu, Meng Jiang, X.~Ren, Clare~R. Voss, and Jiawei Han.
  2018{\natexlab{a}}.
\newblock Automated phrase mining from massive text corpora.
\newblock \emph{IEEE Transactions on Knowledge and Data Engineering},
  30:1825--1837.

\bibitem[{Shang et~al.(2018{\natexlab{b}})Shang, Liu, Ren, Gu, Ren, and
  Han}]{Shang2018LearningNE}
Jingbo Shang, Liyuan Liu, X.~Ren, X.~Gu, Teng Ren, and Jiawei Han.
  2018{\natexlab{b}}.
\newblock Learning named entity tagger using domain-specific dictionary.
\newblock \emph{ArXiv}, abs/1809.03599.

\bibitem[{Thelen and Riloff(2002)}]{Thelen2002ABM}
M.~Thelen and E.~Riloff. 2002.
\newblock A bootstrapping method for learning semantic lexicons using
  extraction pattern contexts.
\newblock In \emph{EMNLP}.

\bibitem[{Yan et~al.(2019)Yan, Han, Sun, and He}]{Yan2019LearningTB}
Lingyong Yan, Xianpei Han, L.~Sun, and B.~He. 2019.
\newblock Learning to bootstrap for entity set expansion.
\newblock In \emph{EMNLP/IJCNLP}.

\bibitem[{Yarowsky(1995)}]{Yarowsky1995UnsupervisedWS}
David Yarowsky. 1995.
\newblock Unsupervised word sense disambiguation rivaling supervised methods.
\newblock In \emph{ACL}.

\bibitem[{Zhang et~al.(2020)Zhang, Shen, Shang, and Han}]{Zhang2020EmpowerES}
Yunyi Zhang, Jiaming Shen, Jingbo Shang, and Jiawei Han. 2020.
\newblock Empower entity set expansion via language model probing.
\newblock In \emph{ACL}.

\end{thebibliography}
\bibliographystyle{acl_natbib}

\clearpage

\appendix
\section{Appendices}

\subsection{Details of Logical Rule Extraction}
\label{app:rules}
In this section, we present details of the extraction and the matching logic of our designed logical rules, using
the following sentence with a location entity \textbf{United States} as an example.
\begin{example}
\label{exp:appendix}
\\
~~\includegraphics[width=0.85\linewidth]{Figures/fig-dep-small-v1.pdf}
\end{example}

We first obtain a parsed dependency tree of the sentence using the spaCy pipeline (\texttt{en\_core\_web\_sm} model). Then our framework will generate all candidate rules for each candidate entity. Here, we use the token span \texttt{United States} as the target candidate entity to show how these rules are extracted.

\smallsection{TokenString} We use the lower-case and lemmatized tokens of an entity candidate as a TokenString rule. Given the above example, we will extract a \mtt{TokenString}=``united state'' rule.

\smallsection{PreNgram} It matches preceding $N$ tokens. All tokens in rules will be lower cased and lemmatized. In our experiments, we set $N$ to $3$. In Example~\ref{exp:appendix}, we extract \mtt{PreNgram}=``the'', 
\mtt{PreNgram}=``to the'', and 
\mtt{PreNgram}=``move to the'' as candidate rules. 

\smallsection{PostNgram} It matches the succeeding $N$ tokens, which are also lower cased and lemmatized. $N$ is set to $3$ in our experiments. In Example~\ref{exp:appendix}, we can extract 
\mtt{PostNgram}=``in'',
\mtt{PostNgram}=``in 1916'', and
\mtt{PostNgram}=``in 1916 .'' as candidate rules.

\smallsection{POSTag} We extract the part-of-speech tags of tokens in a span text using the spaCy pipeline. In Example~\ref{exp:appendix}, we can extract \mtt{POSTag}=``\textsc{propn propn}'' as a candidate rule.

\smallsection{DependencyRel} We first find the head word\footnote{For simplicity, we just used the last token as the head word of a token span.}
in the text span. 
Then, we extract the governor (i.e. head) of the head word as a dependency rule with depth $1$. In Example~\ref{exp:appendix}, \texttt{state} is the head word of text span \texttt{United States}. \texttt{to} is the governor of head word \texttt{state}, so \mtt{DependencyRel}=``to'' is the DependencyRel rule with depth $1$. Next, all tokens dependent on the head word are considered as DependencyRel rules with depth $2$. In Example~\ref{exp:appendix}, word \texttt{move} is logical rule with depth $2$. We use 
$\|$
to connect token with depth $1$ and token with depth $2$. Finally, in Example 1, we have logical rule 
\mtt{DependencyRel}=``to'' and 
\mtt{DependencyRel}=``move$\|$to''.

The numbers of rule candidates for each dataset are: BC5CDR (108,756), CHEMDNER (441, 595), CONLL2003 (142, 976).
\subsection{Details of Neural Tagger}
\label{app:ner_model_detail}
In this section, we present 
details of span representation and prediction in 
our neural tagger.

\smallsection{Span Representation} Given a sentence $\mathbf{x} = [w_1, w_2,\ldots,w_n]$ of $n$ tokens, a span $s_i = [w_{b_i}, w_{b_{i+1}}, \ldots, w_{e_i}]$, where $b_i, e_i$ are the start and end indices respectively. The representation of spans contains two components: a content representation $\mathbf{z}_i^c$ calculated as the weighted average across all token embeddings in the span, and a \textit{boundary representation} $\mathbf{z}_i^u$ that concatenates the embeddings at the start and end positions of the span. Specifically,
\begin{equation*}
    \begin{aligned}
	\mathbf{c}_1, \mathbf{c}_2, \ldots, \mathbf{c}_n &= \mathrm{TokenRepr}(w_1, w_2, \ldots, w_n),\\
	\mathbf{u}_1, \mathbf{u}_2, ..., \mathbf{u}_n &=  \mathrm{BiLSTM}(\mathbf{c}_1, \mathbf{c}_2,\ldots,\mathbf{c}_n),\\
	\mathbf{z}_i^c &= \mathrm{SelfAttn}(\mathbf{c}_{b_i}, \mathbf{c}_{b_i+1}, \ldots, \mathbf{c}_{e_i}),\\
	\mathbf{z}_i^u &= [\mathbf{u}_{b_i}; \mathbf{u}_{e_i}], \mathbf{z}_i = [\mathbf{z}_i^c;\mathbf{z}_i^u],
\end{aligned}
\end{equation*}
where $\mathrm{TokenRepr}$ could be non-contextualized, such as Glove~\cite{Pennington2014GloveGV}, or contextualized, such as BERT~\cite{Devlin2019BERTPO}. $\mathrm{BiLSTM}$ is a bi-directional LSTM layer and $\mathrm{SelfAttn}$ is a self-attention layer. For further details please refer to \citet{Lee2017EndtoendNC}.

\smallsection{Span Prediction} We predict labels for \textit{all} spans up to a fixed length of $l$ words using a multilayer perceptron (MLP):
\begin{equation}
    \mathbf{o}_i = \mathrm{softmax}(\mathbf{MLP}^{\mathrm{span}}(\mathbf{z}_i))
\end{equation}
where $\mathbf{o}_i$ is prediction for the span. We introduce one negative label \texttt{NEG} as an additional label which indicates invalid spans (i.e.,~spans that are not named entities in the corpus).
\begin{figure}[t]
     \centering
     \includegraphics[width=0.48\textwidth]{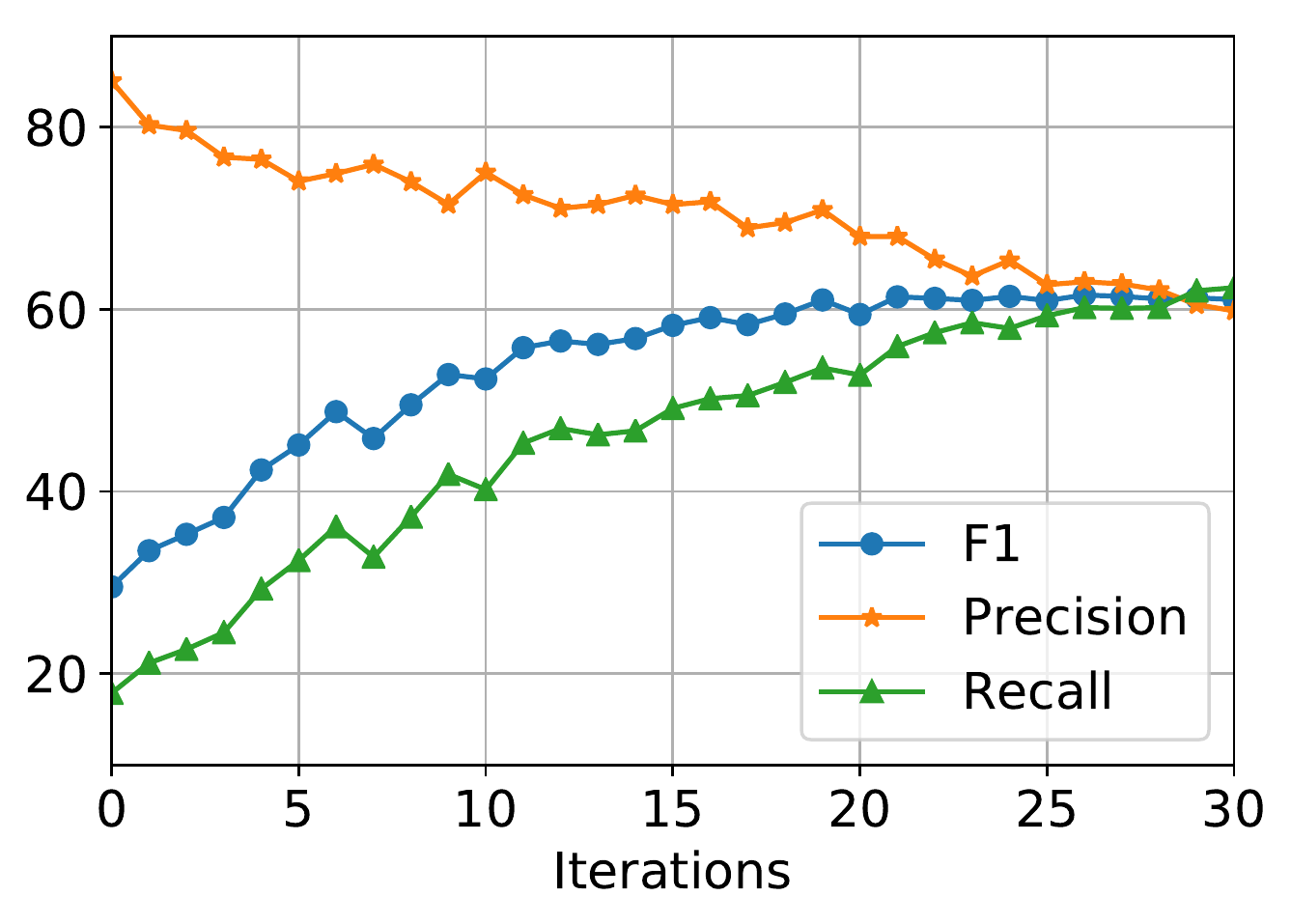}
     \caption{Iterations vs.~performance of the neural NER tagger on CHEMDNER datasets.}
     \label{fig:chemdner_iter}
\end{figure}

\begin{figure}[t]
     \centering
     \includegraphics[width=0.48\textwidth]{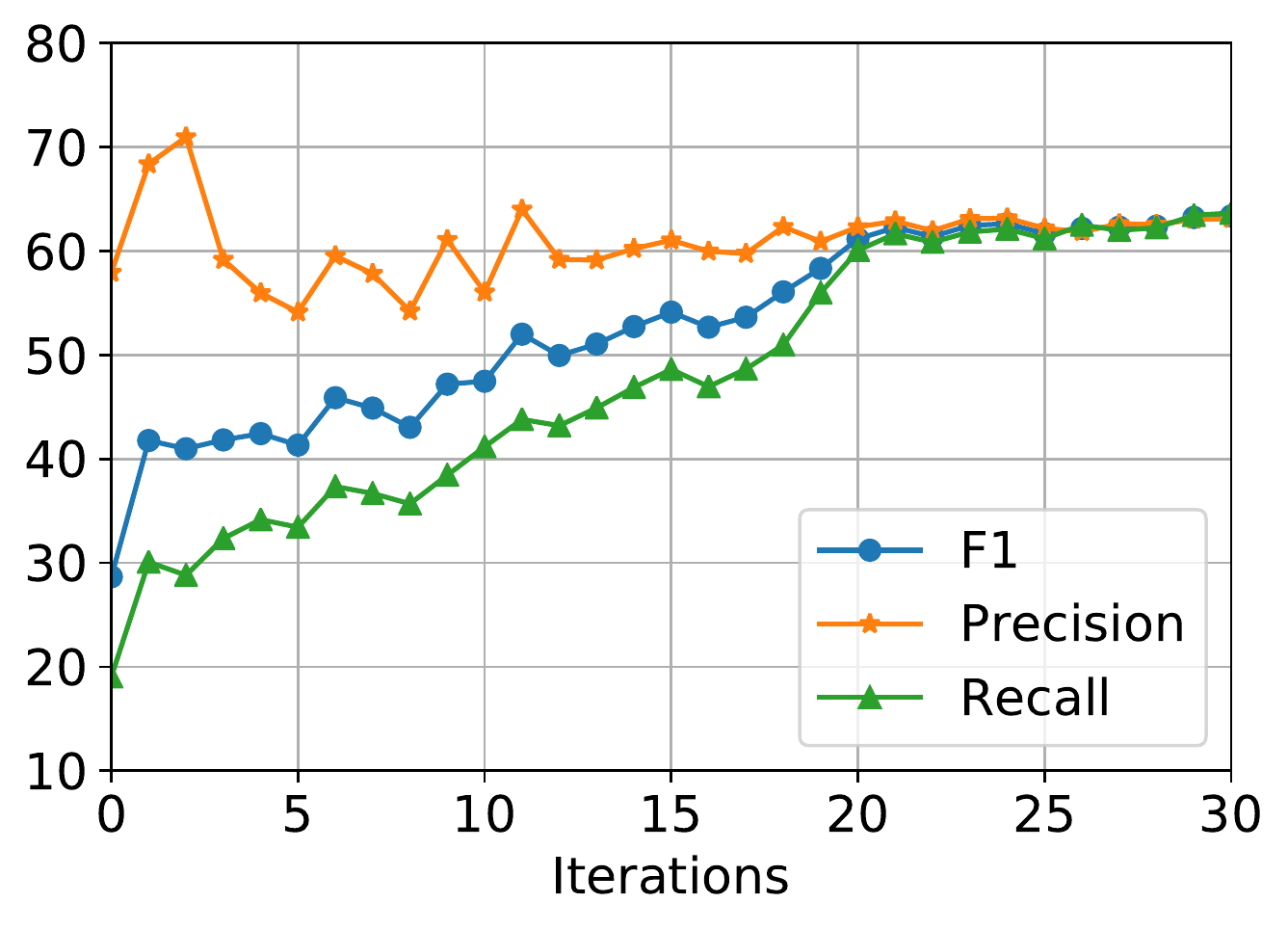}
     \caption{Iterations vs.~performance of the neural NER tagger on CoNLL2003 datasets.}
     \label{fig:conll_iter}
\end{figure}

\begin{table}[h]
    \centering
    \small
    \begin{tabular}{l|c}
        \toprule
        Condition & Label\\      
        \midrule
         \makecell[l]{
         TokenString(x)==``nicotine"\\
         TokenString(x)==``morphine" \\
         TokenString(x)==``haloperidol" \\
         TokenString(x)==``warfarin" \\
         TokenString(x)==``clonidine" \\
         TokenString(x)==``creatinine" \\
         TokenString(x)==``isoproterenol"\\
         TokenString(x)==``cyclophosphamide"\\
         TokenString(x)==``sirolimus" \\
         TokenString(x)==``tacrolimus" 
         }
         
         & Chemical \\   
         \hline
         \makecell[l]{
         TokenString(x)==``proteinuria" \\
         TokenString(x)==``esrd" \\
         TokenString(x)==``thrombosis" \\
         TokenString(x)==``tremor" \\
         TokenString(x)==``hepatotoxicity" \\
         TokenString(x)==``hypertensive" \\
         TokenString(x)==``thrombotic"\\
         TokenString(x)==``microangiopathy"\\
         TokenString(x)==``thrombocytopenia"\\
         TokenString(x)==``akathisia" 
         }
         
         & Disease \\   
        \bottomrule
    \end{tabular}
    \caption{Seed logical rules for BC5CDR dataset.
    }
    \label{app:tab_seed_bc5cdr}
\end{table}

\begin{table}[h]
    \centering
    \small
    \begin{tabular}{l|c}
        \toprule
        Condition & Label\\      
        \midrule
         \makecell[l]{
         TokenString(x)==``britain" \\
         TokenString(x)==``italy" \\
         TokenString(x)==``russia" \\
         TokenString(x)==``sweden“ \\
         TokenString(x)==``belgium” \\
         TokenString(x)==``iraq" \\
         TokenString(x)==``south africa"\\
         TokenString(x)==``united states"
         }
         & Location \\   
         
         \hline
         \makecell[l]{
         TokenString(x)==``wasim akram" \\
         TokenString(x)==``waqar younis" \\
         TokenString(x)==``mushtaq ahmed" \\
         TokenString(x)==``mother teresa"\\
         TokenString(x)==``aamir sohail” \\
         TokenString(x)==``bill clinton" \\
         TokenString(x)==``saeed anwar”
         }
         
         & Person \\ 
         
         \hline
         \makecell[l]{
         TokenString(x)==``osce" \\
         TokenString(x)==``nato" \\
         TokenString(x)==``honda" \\
         TokenString(x)==``interfax" \\
         TokenString(x)==``marseille"
         }
         
         & Organization \\   
         
        \bottomrule
    \end{tabular}
    \caption{Seed logical rules for CoNLL2003 dataset.
    }
    \label{app:tab_seed_conll2003}
\end{table}

\begin{table}[h]
    \centering
    \small
    \begin{tabular}{l|c}
        \toprule
        Condition & Label\\      
        \midrule
         \makecell[l]{
         TokenString(x)==``glucose"\\
         TokenString(x)==``oxygen" \\
         TokenString(x)==``cholesterol" \\
         TokenString(x)==``glutathione" \\
         TokenString(x)==``ethanol" \\
         TokenString(x)==``ca ( 2 + )" \\
         TokenString(x)==``calcium"\\
         TokenString(x)==``androgen" \\
         TokenString(x)==``copper" \\
         TokenString(x)==``graphene" \\
         TokenString(x)==``glutamate" \\
         TokenString(x)==``dopamine" \\
         TokenString(x)==``cocaine" \\
         TokenString(x)==``cadmium" \\
         TokenString(x)==``serotonin"\\
         TokenString(x)==``estrogen" \\
         TokenString(x)==``nicotine" \\
         TokenString(x)==``tyrosine" \\
         TokenString(x)==``resveratrol" \\
         TokenString(x)==``nitric oxide" \\
         TokenString(x)==``cisplatin" \\
         TokenString(x)==``alcohol" \\
         TokenString(x)==``superoxide"\\
         TokenString(x)==``curcumin" \\
         TokenString(x)==``( 1 ) h" \\
         TokenString(x)==``metformin" \\
         TokenString(x)==``amino acid" \\
         TokenString(x)==``arsenic" \\
         TokenString(x)==``zinc" \\
         TokenString(x)==``testosterone" \\
         TokenString(x)==``flavonoids"\\
         TokenString(x)==``camp" \\
         TokenString(x)==``methanol" \\
         TokenString(x)==``amino acids"\\
         TokenString(x)==``mercury" \\
         TokenString(x)==``fatty acids" \\
         TokenString(x)==``polyphenols" \\
         TokenString(x)==``nmda"\\
         TokenString(x)==``silica" \\
         TokenString(x)==``5 - ht"
         }
         & Chemical \\   
        \bottomrule
    \end{tabular}
    \caption{Seed logical rules for CHEMDNER dataset.
    }
    \label{app:tab_seed_chemdner}
\end{table}

\subsection{Negative Instances for Training}
To provide negative supervision for neural network training, we pre-process unlabeled data and collect all noun phrases. Token spans outside noun phrases are used as initial negative supervision. Compared with previous works~\cite{Ratner2017SnorkelRT, Fries2017SwellSharkAG} that directly use noun phrases as entity candidates, in our work, noun phrases only provide negative supervision. In the following iterations, these negative instances still have a chance to be recognized correctly.

\subsection{Parameters}
\label{app:parameters}
In our neural NER tagger, we use the Adam optimizer with learning rate $2e^{-5}$, a dropout ratio $0.5$, and a batch size of $32$ for all experiments. For better stability, we use gradient clipping of $5.0$. In addition, the maximum length of spans is $5$, and precision thresholds for rules are $0.9$ for all experiments. 

In the dynamic label selection step, we set the temperature of thresholds to $0.8$, sample times $N=50$, $E_s=3$, and the temperature $\tau=0.8$ to control threshold.
In logical rule scoring and selection step, we set $\eta=1$, and threshold $\theta=0.9$.

In our experiments, we use SciBert for two biomedical datasets and Bert for CoNLL2003 dataset. During training, we run the framework for $32$ iterations for all datasets and select the best model based on development sets.

\subsection{Implementation}
We implement our framework with Pytorch 1.4.0\footnote{https://pytorch.org/} and our rule labeling is based on Snorkel 0.9.5\footnote{https://www.snorkel.org/}. We train our framework on NVIDIA Quadro RTX 8000 GPU. Our neural NER module has 114,537,220 parameters. It takes about $30$ minutes to complete a whole iteration.

\subsection{Dictionary for AutoNER}
\label{app:dict_autoner}
In Table~\ref{tab:main_perf}, we used the same manual seed rules as supervision for all experiments. For AutoNER, all phrases generated from AutoPhrase are used as untyped phrases (i.e., full dictionary in AutoNER), the sizes are: BC5CDR (6,619), CHEMDNER (15,995), CONLL2003 (4,137). We expanded seeds with CGExpan and used the expansion as the typed terms for AutoNER (i.e., the core dictionary in AutoNER).
We experimented with different sizes of dictionaries and reported the best results.
The sizes for the best performance are: BC5CDR (800), CHEMDNER (500), CONLL2003 (1000).
We found that the performance will be lower when we try to use larger automatically expanded dictionaries. 
\subsection{Seed Logical Rules}
\label{app:seeds}
In this section, we show the seeds used in experiments of Table~\ref{tab:main_perf}.

Seed logical rules for BC5DCR, CoNLL2003 and CHEMDNER is shown in Table~\ref{app:tab_seed_bc5cdr}, \ref{app:tab_seed_conll2003} and \ref{app:tab_seed_chemdner} respectively.

\subsection{Iterations vs.~Performance}
\label{app:perf_iter}
Figure~\ref{fig:chemdner_iter} and Figure~\ref{fig:conll_iter} show the 
performance vs.~iterations on CHEMDNER and CoNLL 2003 dataset.

\end{document}